\pdfoutput=1

\documentclass[11pt]{article}

\usepackage[final]{acl}

\usepackage{times}
\usepackage{latexsym}

\usepackage[T1]{fontenc}

\usepackage[utf8]{inputenc}

\usepackage{microtype}

\usepackage{inconsolata}

\usepackage{graphicx}

\usepackage{amstext}
\usepackage{amsmath}
\usepackage{amssymb}
\usepackage{subcaption}
\usepackage{graphicx}
\usepackage{booktabs}
\usepackage{multirow}
\usepackage{tikz-dependency}
\usepackage{amsthm}
\usepackage{bbm}
\usepackage{thmtools}
\usepackage{thm-restate}
\usepackage{stfloats}

\newtheorem{theorem}{Theorem}

\title{Parallel Continuous Chain-of-Thought with Jacobi Iteration}

\author{Haoyi Wu, Zhihao Teng, Kewei Tu\thanks{\; Corresponding author.} \\
School of Information Science and Technology, ShanghaiTech University \\
Shanghai Engineering Research Center of Intelligent Vision and Imaging \\
\texttt{\{wuhy1, tengzhh2022, tukw\}@shanghaitech.edu.cn}}

\begin{document}
\maketitle
\begin{abstract}

  Continuous chain-of-thought has been shown to be effective in saving reasoning tokens for large language models.
  By reasoning with continuous latent thought tokens, continuous CoT is able to perform implicit reasoning in a compact manner.
  However, the sequential dependencies between latent thought tokens spoil parallel training, leading to long training time.
  In this paper, we propose Parallel Continuous Chain-of-Thought (PCCoT), which performs Jacobi iteration on the latent thought tokens, updating them iteratively in parallel instead of sequentially and thus improving both training and inference efficiency of continuous CoT.
  Experiments demonstrate that by choosing the proper number of iterations, we are able to achieve comparable or even better performance while saving nearly 50\% of the training and inference time.
  Moreover, PCCoT shows better stability and robustness in the training process.
  Our code is available at \url{https://github.com/whyNLP/PCCoT}.

\end{abstract}

\section{Introduction}

Chain-of-thought (CoT) enables large language models (LLMs) to solve complex problems by generating intermediate reasoning steps \cite{wei2023chainofthoughtpromptingelicitsreasoning,chu-etal-2024-navigate,chen2025reasoningerasurveylong}.
However, the explicit nature of CoT can lead to increased token consumption and lower inference speed \cite{sui2025stopoverthinkingsurveyefficient,liu2025efficientinferencelargereasoning}.

Recently, continuous CoT has been shown to be effective in saving reasoning tokens by performing implicit reasoning with continuous vectors (also referred to as latent thought tokens) \cite{hao2024traininglargelanguagemodels,shen2025codicompressingchainofthoughtcontinuous}.
By reasoning in a continuous manner, LLMs have the freedom to reason without being constrained in the discrete language space, thus potentially performing reasoning more compactly and efficiently.
However, because of the sequential dependencies between latent thought tokens that spoil parallel training, existing approaches to continuous CoT suffer from long training time.

In this paper, we propose Parallel Continuous Chain-of-Thought (PCCoT), which performs nonlinear Jacobi iteration \cite{ortega1970iterative} on latent thought tokens to mitigate the above-mentioned issues and improve the efficiency of continuous CoT.
Specifically, we iteratively update all the latent thought tokens in parallel instead of sequentially decoding them.
By choosing the proper numbers of iterations and latent thought tokens, we are able to speed up the reasoning process by a large scale without sacrificing the performance.
Note that PCCoT subsumes previous work as special cases. If performing only a single iteration, then PCCoT is equivalent to Pause Tokens \cite{goyal2024thinkspeaktraininglanguage}.
If the iteration number is equal to the number of latent thought tokens, then PCCoT becomes equivalent to continuous CoT.

Our experiments on math reasoning demonstrate that PCCoT using a small number of iterations could achieve comparable or even better performance than that of continuous CoT with sequential decoding, while saving nearly 50\% of the training and inference time.
Moreover, we observe that PCCoT with small numbers of iterations shows better stability and robustness in the training process.

\begin{figure*}[tb]
  \centering
  \begin{subfigure}{.48\textwidth}
    \centering
    \includegraphics[page=1,width=\textwidth,trim=225 50 225 120,clip]{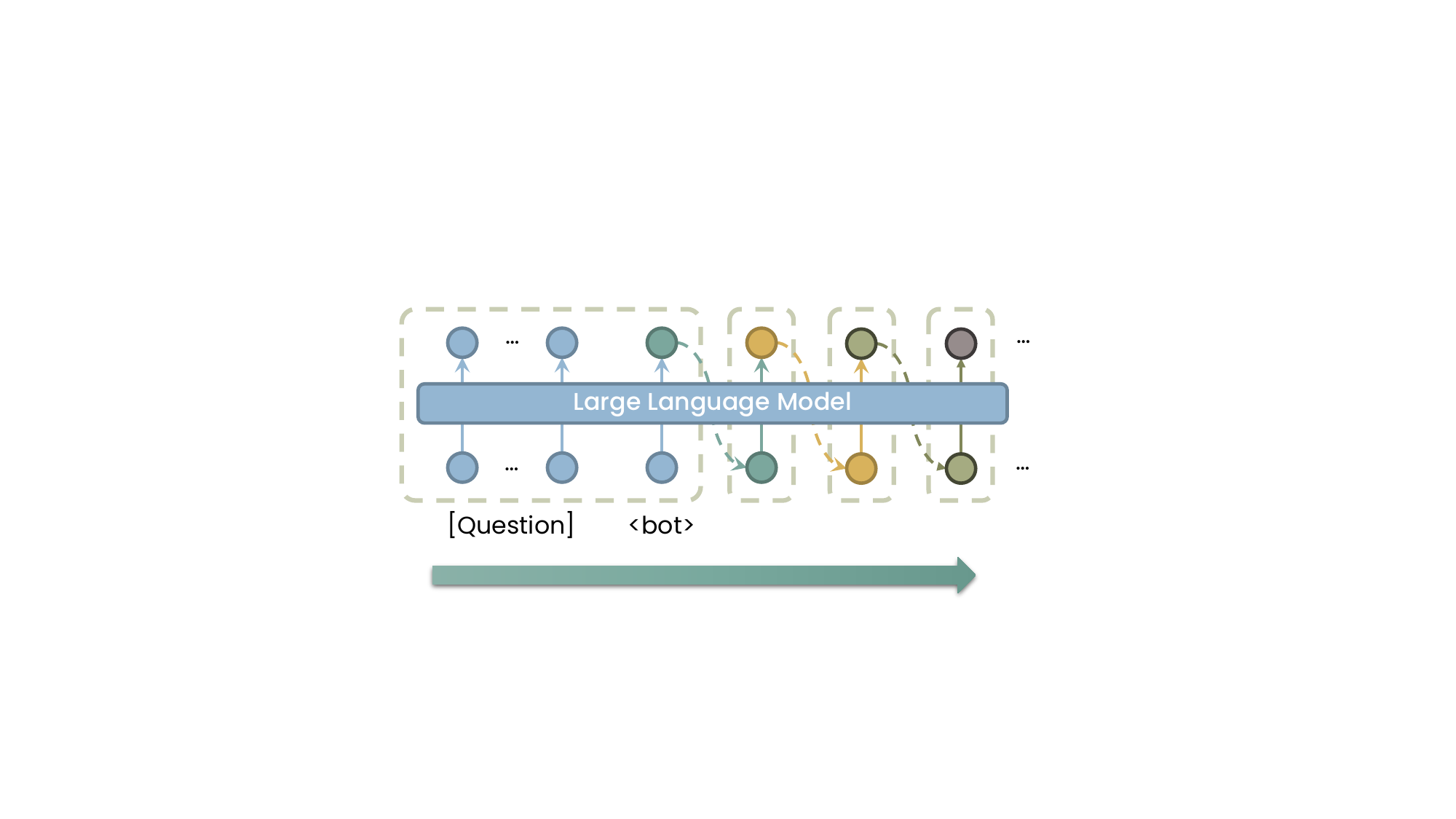}
    \caption{Continuous CoT}
    \label{fig:main-a}
  \end{subfigure}
  \quad
  \begin{subfigure}{.48\textwidth}
    \centering
    \includegraphics[page=2,width=\textwidth,trim=250 50 200 50,clip]{figs/figures.pdf}
    \caption{Parallel Continuous CoT (PCCoT)}
    \label{fig:main-b}
  \end{subfigure}
  \caption{An illustration of Continuous Chain-of-Thought (left) and Parallel Continuous Chain-of-Thought (right). The figure shows $c=3$ latent thought tokens with the first forward pass and $T=2$ extra iterations. The \texttt{<eot>} token and the answer tokens are not shown in the figure. Each dashed box represents a single forward pass.}
  \label{fig:main}
\end{figure*}

\section{Background}

Compared to standard CoT, continuous CoT directly feeds the final hidden state as the input embedding for the next token (i.e., the next latent thought token), instead of mapping the final hidden state to the vocabulary and then embedding the selected next token to the hidden space to form the next input vector. Figure~\ref{fig:main-a} shows an illustration of continuous CoT.
We follow the paradigm in Coconut \cite{hao2024traininglargelanguagemodels} and formally define continuous CoT as follows.

Let $x = (x_1, x_2, \ldots, x_n)$ be the query sequence.
Continuous CoT first appends a learnable special token $x_{n+1}=\texttt{<bot>}$ representing the beginning of thought to the input sequence and feeds it to the transformer model.
The computation of latent thought tokens is as follows:
\begin{align*}
  h_{n+1} &= f([E_{x_{1}}; \dots; E_{x_{n+1}}]) \\
  h_{n+i+1} &= f([E_{x_{1}}; \dots; E_{x_{n+1}}; h_{n+1}; \dots; h_{n+i}])
\end{align*}
where $i = 1, 2, \dots, c$, $c$ is the number of latent thought tokens, $f$ is the transformer model without the prediction head, $E_{x_{j}}$ is the embedding vector of token $x_{j}$. $[\cdot; \cdot]$ represents the concatenation of two (or more) vectors.

After generating latent thought tokens, continuous CoT appends the end-of-thought token $x_{n+c+2}=\texttt{<eot>}$ to the input sequence and then generates the answer tokens sequentially in the same way as the standard transformer.

\section{Parallel Continuous Chain-of-Thought}

\subsection{Jacobi Iteration}

Because of the sequential dependencies between latent thought tokens, existing approaches to continuous CoT cannot perform parallel training, leading to long training time.
To this end, we propose to perform Jacobi iteration on the latent thought tokens to improve the efficiency of continuous CoT, which we refer to as Parallel Continuous Chain-of-Thought (PCCoT). Figure~\ref{fig:main-b} shows an illustration of PCCoT.

Instead of decoding the latent thought tokens sequentially, we iteratively update all the latent thought tokens in parallel.
Given the query sequence $x = (x_1, x_2, \ldots, x_n)$, we first append the begin-of-thought token $x_{n+1}=\texttt{<bot>}$ and $c$ dummy latent thought tokens $x_{n+i+1} = \texttt{<latent>} (i = 1, 2, \cdots, c)$ to the input sequence, and then feed it to the transformer model:
\begin{equation*}
    [h^{(1)}_{n+1}; \dots; h^{(1)}_{n+c+1}] = f([E_{x_{1}}; \dots; E_{x_{n+c+1}}])
\end{equation*}
For the next $T$ extra iterations, we update the input vectors of the latent thought tokens as the final hidden state vectors of the previous token in the last iteration:
\begin{align*}
    [h^{(t+1)}_{n+1}; \dots; h^{(t+1)}_{n+c+1}] &= f([E_{x_{1}}; \dots; E_{x_{n+1}}; \\
    & h^{(t)}_{n+1}; \dots; h^{(t)}_{n+c}])
\end{align*}
where $t=1,2,\dots,T$ and $T$ is the number of extra iterations.
After $T$ extra iterations, we append the end-of-thought token $x_{n+c+2}=\texttt{<eot>}$ to the input sequence and generate the answer tokens based on the hidden states computed from the last iteration in the same way as the standard transformer.

\subsection{Relation to Other Methods}
\label{sec:method-relation}

PCCoT is closely related to a few existing approaches. In fact, with different settings of the number of continuous thought tokens $c$ and the number of extra iterations $T$, PCCoT can be reduced to these existing approaches.

\paragraph{Implicit Chain-of-Thought (iCoT)}
Implicit Chain-of-Thought (iCoT) \cite{deng2024explicitcotimplicitcot} removes all reasoning tokens and directly decodes the answer tokens.
By setting $c=0$, PCCoT is nearly equivalent to iCoT.

\paragraph{Pause Tokens}
Pause Tokens \cite{goyal2024thinkspeaktraininglanguage} appends trainable discrete tokens to the input sequence to allow new computational pathways.
By setting $c>0$ but not performing any extra iterations ($T=0$), PCCoT is nearly equivalent to Pause Tokens.

\paragraph{Continuous Chain-of-Thought}
The only difference between PCCoT and continuous CoT \cite{hao2024traininglargelanguagemodels} is that PCCoT performs Jacobi iteration on the latent thought tokens.
It can be proved that with a sufficient number of iterations, the computation graph of PCCoT is equivalent to that of continuous CoT.
We leave the formal proof to Appendix~\ref{apx:relation-proof}.

\subsection{Training Method}
\label{sec:method-objective}
PCCoT does not require any specialized training procedure and is compatible with any existing training methods of continuous CoT.
In this work, we adopt CODI \cite{shen2025codicompressingchainofthoughtcontinuous} for training as it has the best performance in the literature.
Specifically, CODI jointly trains a teacher task and a student task with a shared model.
The teacher task learns standard CoT with the standard cross-entropy loss on gold reasoning and answer tokens.
The student task learns continuous CoT with the cross-entropy loss on the answer tokens only.
CODI additionally distills the knowledge from the teacher task to the student task by minimizing the L1 loss between the teacher and student prediction distributions on the last token of the answer prompt (``:'' in ``The answer is:'').
CODI uses a MLP to enhance the hidden representation of the latent thought tokens, but we do not use it for a fair comparison with the baseline and other methods.

\section{Experiments}

\begin{table}[tb]
  \centering
  \begin{tabular}{@{}lll@{}}
  \toprule
  Approach          & GSM8K                             & GSM8K-NL                           \\ \midrule
  \multicolumn{3}{l}{\textbf{GPT-2 Small}}                                                   \\
  CoT              & 44.1                              & 34.8                               \\
  Implicit CoT     & 37.78 {\small $\pm 0.31$}          & 37.72 {\small $\pm 1.10$}          \\
  Pause Tokens     & 39.27 {\small $\pm 0.46$}          & 33.79 {\small $\pm 2.10$}          \\
  Continuous CoT   & 48.24 {\small $\pm 1.61$}          & 45.06 {\small $\pm 2.58$}          \\
  PCCoT (Ours)      & \textbf{49.48 {\small $\pm 0.31$}} & \textbf{49.23 {\small $\pm 0.80$}} \\ \midrule
  \textit{CODI}    & 43.7                               & 35.3                               \\
  \textit{Coconut} & 34.1 {\small $\pm 1.5$}            & --                                 \\
  \textit{iCoT}    & 30                                 & 3.2                                \\ \midrule
  \multicolumn{3}{l}{\textbf{Llama3.2-1B-Instruct}}                                          \\
  CoT              & 61.6                              & 54.1                               \\
  Implicit CoT     & 52.36 {\small $\pm 0.74$}          & 47.89 {\small $\pm 0.89$}          \\
  Pause Tokens     & 51.78 {\small $\pm 0.91$}          & 48.07 {\small $\pm 0.73$}          \\
  Continuous CoT   & 50.47 {\small $\pm 0.68$}          & 48.47 {\small $\pm 1.40$}          \\
  PCCoT (Ours)      & \textbf{53.35 {\small $\pm 0.18$}} & \textbf{50.72 {\small $\pm 1.39$}} \\ \midrule
  \textit{CODI}    & 55.6                               & 49.7                               \\ \bottomrule
  \end{tabular}
  \caption{
    Test set accuracy (\%) of different methods on GSM8K-Aug and GSM8K-Aug-NL.
    The results of Implicit CoT, Pause Tokens, Continuous CoT and PCCoT are averaged over 3 random runs with standard deviations also shown.
    The results of CoT, CODI \cite{shen2025codicompressingchainofthoughtcontinuous}, Coconut \cite{hao2024traininglargelanguagemodels} and iCoT \cite{deng2024explicitcotimplicitcot} are taken from the literature.
  }
  \label{tab:main}
\end{table}

\subsection{Setup}
\label{sec:exp-setup}

Following \citet{shen2025codicompressingchainofthoughtcontinuous}, we use GSM8K-Aug and GSM8K-Aug-NL \cite{deng2023implicitchainthoughtreasoning} as our datasets.
These datasets are extended from GSM8K \cite{cobbe2021trainingverifierssolvemath} with a 385k training set while leaving the test set unchanged.
GSM8K-Aug uses only math equations as the reasoning steps, while GSM8K-Aug-NL uses natural language as the reasoning steps.

We use the pretrianed GPT-2 \cite{radford2019language} and Llama3.2-1B-Instruct \cite{grattafiori2024llama3herdmodels} as our base models and apply LoRA \cite{hu2022lora} to fine-tune the models.
We mainly follow the hyperparameters in \citet{shen2025codicompressingchainofthoughtcontinuous}.
We use a batch size of 128, a LoRA rank of $r=128$ and a LoRA alpha value of 32.
We use the AdamW \cite{loshchilov2018decoupled} optimizer with a learning rate of $3\times 10^{-3}$ and weight decay of $0.01$ for GPT-2, and a learning rate of $8\times 10^{-4}$ and weight decay of $0.1$ for Llama3.2-1B-Instruct.
We train GPT-2 for 40 epochs and Llama3.2-1B-Instruct for 10 epochs.
More details can be found in Appendix~\ref{apx:exp}.

We compare PCCoT with Implicit CoT, Pause Tokens and Continuous CoT, all trained with the method in Section~\ref{sec:method-objective}.
Pause Tokens uses 24 trainable pause tokens, continuous CoT uses 12 latent thought tokens, and PCCoT uses $c=24$ latent thought tokens with $T=3$ extra iterations.
We also compare with the state-of-the-art results reported in the literature, including CODI \cite{shen2025codicompressingchainofthoughtcontinuous}, Coconut \cite{hao2024traininglargelanguagemodels} and iCoT \cite{deng2024explicitcotimplicitcot}.

\subsection{Results}
\label{sec:exp-results}

\begin{table}[tb]
  \begin{tabular}{@{}lcc@{}}
  \toprule
  Approach        & Training (h) & Inference (s) \\ \midrule
  CoT            & 4.39         & 1.353         \\
  Implicit CoT   & 11.23        & 0.040         \\
  Pause Tokens   & 11.88        & 0.041         \\
  Continuous CoT & 24.91        & 0.443         \\
  PCCoT (Ours)    & 13.72        & 0.199         \\ \bottomrule
  \end{tabular}
  \caption{Training and inference time of different methods with GPT-2 Small on GSM8K-Aug. The inference time is measured with a batch size of 100 and only the time for processing the question and CoT tokens is included.}
  \label{tab:latency}
\end{table}

We report the average and standard deviation of the test set accuracy from 3 random runs in Table~\ref{tab:main}.
The CoT baseline is taken from CODI \cite{shen2025codicompressingchainofthoughtcontinuous}.

Compared with baseline methods, PCCoT achieves the best performance on both datasets with smaller standard deviation in most cases.
This indicates that PCCoT not only acquires better reasoning ability, but is also more robust and stable during training.

Table~\ref{tab:latency} shows the training and inference time of different methods with GPT-2 Small on GSM8K-Aug.
We use two H800 GPUs for training and use one A6000 GPU for inference.
The inference time is measured with a batch size of 100 and only the time for processing the question and CoT tokens is included.
We can see that PCCoT achieves a significant speedup in both training and inference compared with continuous CoT while achieving better performance.

\begin{figure}[tb]
  \centering
  \includegraphics[page=1,width=0.48\textwidth,trim=0 0 0 0,clip]{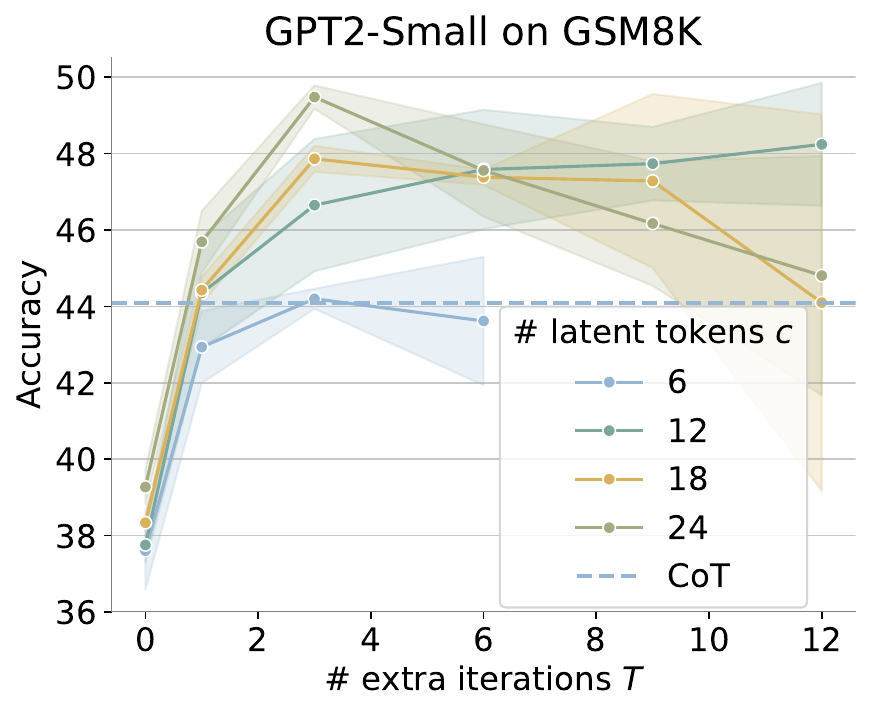}
  \caption{Test set accuracy (\%) of PCCoT with different number of extra iterations $T$ and latent thought tokens $c$ on GSM8K-Aug. The figure shows the average over 3 random runs with standard deviation.}
  \label{fig:exp-iter-token}
\end{figure}

We further plot test set accuracy of PCCoT with different numbers of extra iterations $T$ and latent thought tokens $c$ on GSM8K-Aug (Figure~\ref{fig:exp-iter-token}).
Interestingly, we find that increasing the number of iterations does not necessarily improve the performance.
As $T$ increases, the model performance first has a significant improvement at about $T=3$ and then starts to decrease (except for $c=12$ latent tokens).
Moreover, with a large number of $T$, training becomes unstable, which leads to a large standard deviation.

\section{Related Work}

\subsection{Continuous Chain-of-Thought}

Previous studies have explored several approaches to training models with continuous CoT.
iCoT \cite{deng2024explicitcotimplicitcot} and COCONUT \cite{hao2024traininglargelanguagemodels} utilize special training curricula to help models learn to reason in latent space.
Reasoning with Latent Thoughts \cite{saunshi2025reasoninglatentthoughtspower} and RELAY \cite{yu2025enhancingautoregressivechainofthoughtloopaligned} illustrate that looped transformers naturally induce latent thoughts at each iteration.
CCoT \cite{cheng2024compressedchainthoughtefficient} and SoftCoT \cite{xu2025softcotsoftchainofthoughtefficient} train auxiliary models to generate contentful continuous CoT tokens to help generate answers.
Implicit-KD \cite{deng2023implicitchainthoughtreasoning} introduces a knowledge distillation paradigm where a teacher model generates explicit CoT tokens and a student model learns to generate continuous CoT tokens.
CODI \cite{shen2025codicompressingchainofthoughtcontinuous} proposes a novel self-distillation framework, where a single model acts as both the teacher and the student.

\subsection{Jacobi Decoding}
Jacobi decoding \cite{santilli-etal-2023-accelerating} applies the Jacobi iteration method to the decoding process of autoregressive language models.
There have been extensive researches on Jacobi decoding accelerating transformers in terms of inference.
Lookahead decoding \cite{lookahead} improves the efficiency of Jacobi decoding by leveraging n-grams generated from previous Jacobi iterations.
CLLM \cite{kou2024cllms} proposes a training approach specialized for Jacobi decoding that greatly improves the efficiency of the Jacobi decoding process.
Jacobi iteration in PCCoT differs from that in Jacobi decoding in that PCCoT does not involve any discrete token decoding, thus enabling the use of Jacobi iteration in both training and inference.


\section{Conclusion}

In this paper, we propose Parallel Continuous Chain-of-Thought (PCCoT), which performs Jacobi iteration on latent thought tokens to improve the efficiency of continuous CoT.
Experiments demonstrate that PCCoT with a small number of iterations could achieve comparable or even better performance than that of continuous CoT, while saving nearly 50\% of the training and inference time.
PCCoT also shows better stability and robustness in training and hence is more reliable than continuous CoT.

\section*{Limitations}


The current training method of PCCoT (i.e., CODI) relies on distillation from the CoT teacher task.
Though PCCoT is much faster than continuous CoT, it is still much slower than the standard CoT in terms of training due to the distillation training strategy.

We have analyzed the behavior of PCCoT on latent thought tokens but fail to explain some of the findings.
For example, we find that the latent tokens do not converge in multiple iterations after training, which is not a problem for CoT inference but is nonetheless counter-intuitive.
Before scaling up PCCoT to larger models and more diverse settings, it is necessary to figure out how PCCoT works and what the latent thought tokens are doing. See Appendix~\ref{apx:analysis} for more details.

\section*{Acknowledgements}
This work was supported by the robotic AI-Scientist platform of Chinese Academy of Science, the HPC platform of ShanghaiTech University, and the Core Facility Platform of Computer Science and Communication, SIST, ShanghaiTech University.
We also would like to thank Zongru Liu for his assistance in running part of the experiments.

\bibliography{anthology,custom}

\appendix

\section{Relation to Continuous CoT}
\label{apx:relation-proof}

In Section~\ref{sec:method-relation}, we mention that the computation graph of PCCoT is equivalent to that of continuous CoT with $c$ latent thought tokens with sufficient number of iterations. In this section, we provide a formal proof of this statement.

\begin{theorem}
  \label{thm:equivalent}
  The computation graph of PCCoT with $c$ latent thought tokens and $T$ extra iterations is equivalent to that of continuous CoT with $c$ latent thought tokens if $T \geq c$.
\end{theorem}

The proof is straightforward. After the first iteration, the hidden states of all layers of the first $n+1$ tokens in PCCoT are the same as that in continuous CoT.
By mathematical induction, after the $i$th extra iteration, the hidden states of the $i$th latent thought token are the same as that in continuous CoT.
Therefore, with $T \geq c$ extra iterations, all the latent thought tokens are updated to the same hidden states in both cases.
We formally prove this statement in the following.

\begin{proof}
We prove the theorem by mathematical induction.

\paragraph{Base case}
First, consider $t=1$. Since
\begin{equation*}
  [h^{(1)}_{n+1}; \dots; h^{(1)}_{n+c+1}] = f([E_{x_{1}}; \dots; E_{x_{n+c+1}}])
\end{equation*}
we have
\begin{equation*}
  h^{(1)}_{n+1} = f([E_{x_{1}}; \dots; E_{x_{n+1}}])
\end{equation*}
also
\begin{equation*}
  h_{n+1} = f([E_{x_{1}}; \dots; E_{x_{n+1}}])
\end{equation*}
we have $h^{(1)}_{n+1} = h_{n+1}$.

\paragraph{Inductive step}
We assume that $h^{(t)}_{n+i} = h_{n+i}, \forall t \geq i$ for $i=1,2,\dots,k$, where $k \leq c$.
Consider $k+1$.

Since
\begin{align*}
  [h^{(t+1)}_{n+1}; \dots; h^{(t+1)}_{n+c+1}] &= f([E_{x_{1}}; \dots; E_{x_{n+1}}; \\
    & h^{(t)}_{n+1}; \dots; h^{(t)}_{n+c}])
\end{align*}
we have
\begin{align*}
  [h^{(t+1)}_{n+1}; \dots; h^{(t+1)}_{n+k+1}] &= f([E_{x_{1}}; \dots; E_{x_{n+1}}; \\
    & h^{(t)}_{n+1}; \dots; h^{(t)}_{n+k}])
\end{align*}
and $h^{(t)}_{n+i} = h_{n+i}, \forall t \geq i, i=1,2,\dots,k$, we have
\begin{align*}
  [h^{(t+1)}_{n+1}; \dots; h^{(t+1)}_{n+k+1}] &= f([E_{x_{1}}; \dots; E_{x_{n+1}}; \\
    & h_{n+1}; \dots; h_{n+k}])
\end{align*}
$\forall t \geq k$, i.e.
\begin{align*}
  h^{(t+1)}_{n+1} &= f([E_{x_{1}}; \dots; E_{x_{n+1}}]) \\
  h^{(t+1)}_{n+2} &= f([E_{x_{1}}; \dots; E_{x_{n+1}}; h_{n+1}]) \\
  &\cdots \\
  h^{(t+1)}_{n+k+1} &= f([E_{x_{1}}; \dots; E_{x_{n+1}}; h_{n+1}; \dots; h_{n+k}])
\end{align*}
Also,
\begin{align*}
  h_{n+1} &= f([E_{x_{1}}; \dots; E_{x_{n+1}}]) \\
  h_{n+2} &= f([E_{x_{1}}; \dots; E_{x_{n+1}}; h_{n+1}]) \\
  &\cdots \\
  h_{n+k+1} &= f([E_{x_{1}}; \dots; E_{x_{n+1}}; h_{n+1}; \dots; h_{n+k}])
\end{align*}
Therefore,
\begin{align*}
  h^{(t+1)}_{n+1} &= h_{n+1} \\
  h^{(t+1)}_{n+2} &= h_{n+2} \\
  &\cdots \\
  h^{(t+1)}_{n+k+1} &= h_{n+k+1}
\end{align*}
$\forall t \geq k$.

i.e. $h^{(t)}_{n+i} = h_{n+i}, \forall t \geq i$ for $i=1,2,\dots,k+1$.

Thus, by induction, we have $h^{(t)}_{n+i} = h_{n+i}, \forall t \geq i$ for $i=1,2,\dots,c+1$. Therefore, if $T \geq c$, then $T+1 \geq c+1$, we have $h^{(T+1)}_{n+i} = h_{n+i}, \forall i=1,2,\dots,c+1$. All the latent thought tokens in PCCoT have the same vector representation as those in continuous CoT. The computation graph of PCCoT is equivalent to that of continuous CoT.
\end{proof}

\section{Analysis}
\label{apx:analysis}

\subsection{More Results}

\begin{figure}[tb]
  \centering
  \includegraphics[page=1,width=0.48\textwidth,trim=0 0 0 0,clip]{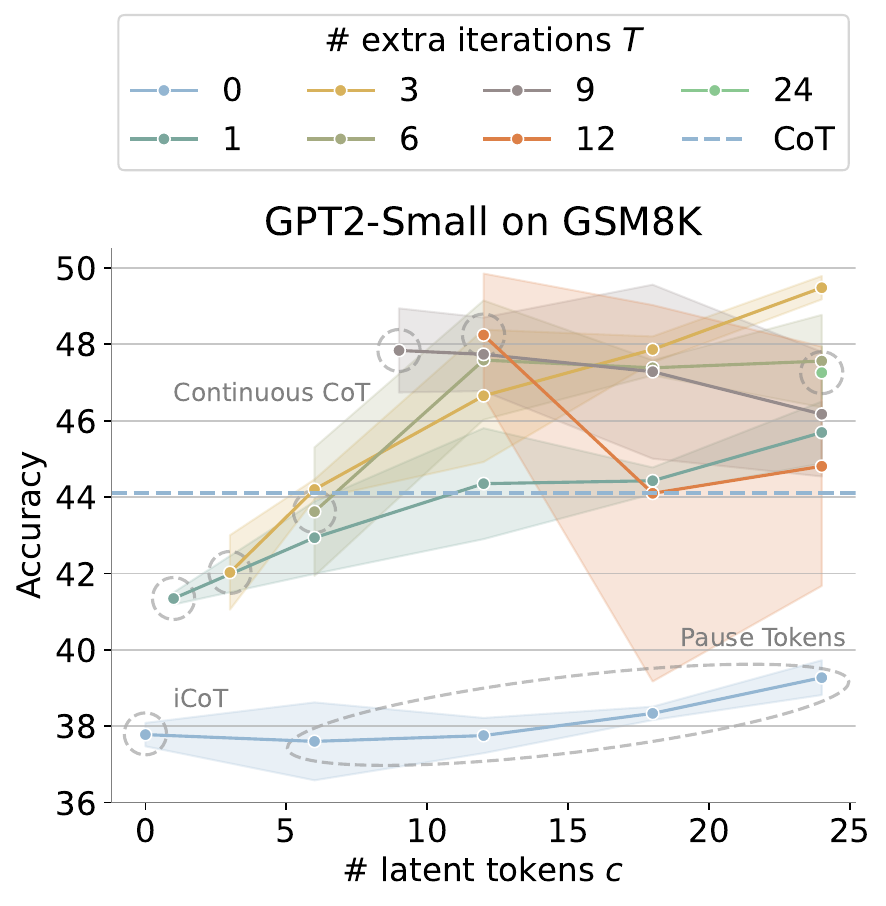}
  \caption{Test set accuracy (\%) of PCCoT with different latent thought tokens $c$ and number of extra iterations $T$ on GSM8K-Aug. The figure shows the average over 3 random runs with standard deviation.}
  \label{fig:exp-token-iter}
\end{figure}

In Section~\ref{sec:exp-results}, we show the test set accuracy of PCCoT with different number of extra iterations $T$ and latent thought tokens $c$ on GSM8K-Aug.
In Figure~\ref{fig:exp-token-iter}, we switch the x-axis and legend of Figure~\ref{fig:exp-iter-token} to show how the number of latent thought tokens $c$ affect the performance of PCCoT.
We also annotate the settings that corresponds to iCoT (no latent thought tokens), Pause Tokens (no extra iterations) and continuous CoT (the number of extra iterations $T$ is equal to the number of latent thought tokens $c$).

It can be seen that with $T = 0, 1, 3$, the performance of PCCoT is stably improved with the increase of the number of latent thought tokens $c$.
The standard deviation of the performance is also small, which indicates that the training process is stable and robust.
It is worth noting that even with $T=1$ extra iteration (the green line), PCCoT outperforms Pause Tokens (the blue line) by a large margin on any number of latent thought tokens $c$.
With $T=3$ extra iterations (the yellow line), the performance of PCCoT increases even faster with the number of latent thought tokens $c$.
However, with over $T=6$ extra iterations, the model performance does not show stable improvement and starts to fluctuate heavily.
This may explain why continuous CoT cannot scale up the number of latent thought tokens \cite{hao2024traininglargelanguagemodels,shen2025codicompressingchainofthoughtcontinuous}, as using too many latent thought tokens would require a large number of iterations (forward passes).
This leads to instability in the training process and limits the performance of the model.

\subsection{Convergence of Latent Thought Tokens}

In Theorem~\ref{thm:equivalent}, we can conclude that the latent thought tokens in PCCoT will eventally converge at $t = c$ extra iterations.
We thus attempt to inspect how the latent thought tokens converge during these iteration updates.

\begin{figure}[tb]
  \centering
  \includegraphics[page=1,width=0.48\textwidth,trim=0 0 0 0,clip]{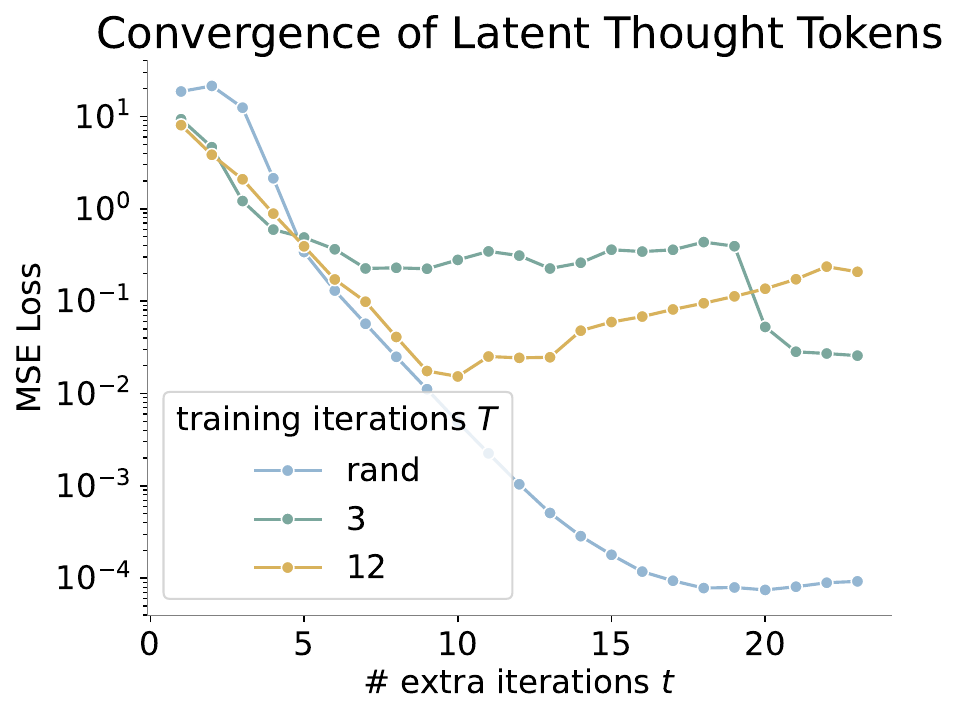}
  \caption{MSE of the latent thought tokens before and after the $t$th extra iteration. ``rand'' means the model is randomly initialized. Other models are trained with $c=24$ and different $T$. The model is tested on random samples from the test set of GSM8K.}
  \label{fig:exp-convergence}
\end{figure}

Following \citet{wu-tu-2024-layer}, we measure the change of the latent thought tokens over consecutive iterations using the mean squared error.
Figure~\ref{fig:exp-convergence} shows the convergence of latent thought tokens of a randomly initialized model, a PCCoT model trained with $c=24, T=3$ and a PCCoT model trained with $c=24, T=12$.
We set the number of latent thought tokens $c=24$ when testing.
The input is a randomly selected batch with size 64 from the test set of GSM8K.
We measure the change of the latent thought tokens over consecutive iterations using the mean squared error (MSE).
Since Theorem~\ref{thm:equivalent} has proved that the $i$th latent thought token will reach a fixed point after $i$ extra iterations, we exclude the tokens that have reached the fixed point from the MSE computation.

From Figure~\ref{fig:exp-convergence}, we can see that a randomly initialized model would perfectly converge as the number of iteration increases.
However, no matter what $T$ is, after training the latent thought tokens no longer converge and stays fluctuating after a certain number of iterations.
This may indicate that Jacobi iterations may not operate as we expect: The success of PCCoT is not due to the fast convergence of the latent thought tokens.

\subsection{Similarities between Latent Thought Tokens}

\begin{figure*}[tb]
  \centering
  \includegraphics[page=1,width=0.9\textwidth,trim=0 0 0 0,clip]{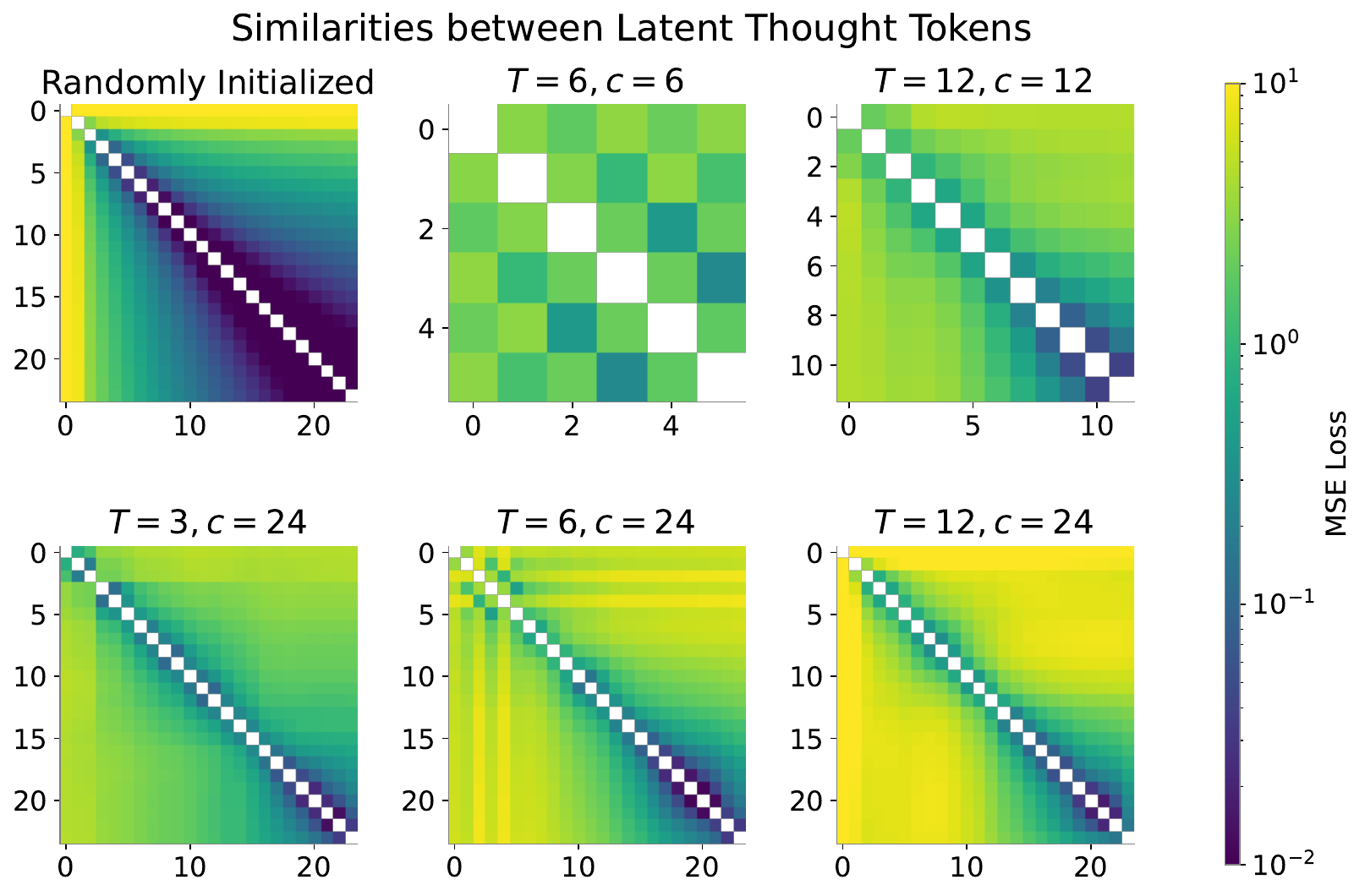}
  \caption{MSE between the latent thought tokens. The darker the block is, the more similar the latent thought tokens are. The model is tested on random samples from the test set of GSM8K.}
  \label{fig:exp-convergence}
\end{figure*}

We also inspect the similarity between the latent thought tokens.
Figure~\ref{fig:exp-convergence} shows the MSE between the latent thought tokens of a randomly initialized model and some PCCoT models.
The axis shows the indices of the latent thought tokens, and the color of the block indicates the MSE between the two latent thought tokens.
The darker the block is, the more similar the latent thought tokens are.

It can be seen that with larger token indices, the latent thought tokens are more similar to each other.
Compared to the randomly initialized model, the latent thought tokens of the PCCoT models are less similar to each other.
With $T=3, c=24$, we can clearly find that the first three latent thought tokens are less similar to other latent thought tokens.

Interestingly, with $T=6$, the first six latent thought tokens show an interleaved pattern.
The latent thought tokens with odd indices are more similar to each other, but less similar to the latent thought tokens with even indices.
Also, the latent thought tokens with even indices are more similar to each other, but less similar to the latent thought tokens with odd indices.
This pattern is also observed by further increasing the number of iterations in $T=3$ settings, but is not clear in $T=12$ settings.
In randomly initialized models, this interleaved pattern is not observed.
Up to now, we have not found a clear explanation for this phenomenon.
Perhaps this indicates that the latent thought tokens have some interdependencies, but we still do not know what does this mean and how it may potentially affect PCCoT in terms of scaling up and extending to more general and complex tasks.

\subsection{The Initialization of Latent Thought Tokens}

\begin{table}[tb]
  \centering
  \begin{tabular}{@{}ll@{}}
  \toprule
  Approach               & Accuracy                  \\ \midrule
  PCCoT                 & 49.48 {\small $\pm 0.31$} \\
  + untrainable latent tokens & 47.31 {\small $\pm 1.07$} \\ \bottomrule
  \end{tabular}
  \caption{Test set accuracy (\%) of PCCoT and its variant with untrainable latent thought tokens on GSM8K-Aug.}
  \label{tab:fixed}
\end{table}

In this section, we inspect how dependent latent thought tokens are on the initialization (i.e. the embedding of \texttt{<latent>} token).
Instead of making the \texttt{<latent>} token trainable, we freeze the starting state of all the latent thought tokens to the same random initialization during training.
i.e. $E_{\texttt{<latent>}}$ is a fixed vector, randomly initialized and not trainable.

Table~\ref{tab:fixed} shows the test set accuracy of PCCoT and its variant with fixed latent thought tokens on GSM8K-Aug.
We use $T=3$ extra iterations and $c=24$ latent thought tokens.
Although the latent thought tokens are fixed to a random initialization, the model still achieves a reasonable performance.
This indicates that the latent thought tokens are robust to the initialization and the model can still learn to update and use them as reasoning tokens.

Note that this does not mean that $E_{\texttt{<latent>}}$ is not important to the model.
Instead, if we perturb $E_{\texttt{<latent>}}$ on a trained model by changing them to a random vector, the accuracy drops to 0.0\%.

\subsection{Comparison with Standard CoT}

\begin{table*}[tb]
  \centering
  \begin{tabular}{@{}lll@{}}
  \toprule
  Approach               & GSM8K & GSM8K-NL \\ \midrule
  CoT                   & 44.1  & 34.8     \\
  PCCoT                 & 49.48 & 49.23    \\
  PCCoT w/ CoT decoding & 50.42 & 39.80    \\
  CoT w/ gold reasoning steps & 88.21 & 87.95    \\
  PCCoT w/ CoT decoding \& gold reasoning steps & 62.24 & 55.88    \\ \bottomrule
  \end{tabular}
  \caption{Test set accuracy (\%) on GSM8K and GSM8K-NL. We compare the performance of PCCoT with standard CoT and PCCoT with standard CoT decoding. Models are finetuned from GPT-2 Small.}
  \label{tab:cmp-cot}
\end{table*}

From Table~\ref{tab:main}, Figure~\ref{fig:exp-iter-token} and Figure~\ref{fig:exp-token-iter}, we can see that the performance of PCCoT surpasses that of standard CoT on GPT-2 Small, especially on GSM8K-Aug-NL.
Since we adopt the distillation training method (CODI) that the student distills the knowledge from the teacher CoT, we also compare the performance of PCCoT with standard CoT decoding, which means that we use the weights of the PCCoT model but decode the reasoning tokens with standard CoT decoding.
Table~\ref{tab:cmp-cot} shows the test set accuracy on GSM8K and GSM8K-NL.

There are two counter-intuitive observations in Table~\ref{tab:cmp-cot}.
\begin{itemize}
  \item The performance of PCCoT is better than that of PCCoT with standard CoT decoding on GSM8K-NL.
  \item The performance of PCCoT with standard CoT decoding is better than that of standard CoT.
\end{itemize}

For the first observation, note that we adopt the distillation training method that the student distills the knowledge from the teacher, thus it is counter-intuitive that the performance of PCCoT (the student task) is better than that of PCCoT with standard CoT decoding (the teacher task).

One possible explanation is that the standard CoT has a gap between training and inference: during training, the model learns to generate the next token based on the gold previous tokens, while during inference, the model generates tokens autoregressively and if the model makes a mistake, it will propagate to the next token.
In PCCoT, since the latent thought tokens are continuous vectors, such a gap does not exist in the reasoning process.
Therefore, on GSM8K-Aug-NL, the reasoning path is much longer and it is more likely that the standard CoT model will make mistakes. However, PCCoT avoids this problem in reasoning and thus its performance surpasses that of standard CoT.

To verify this, we evaluate the performance of CoT with gold reasoning steps and PCCoT with gold reasoning steps and standard CoT decoding.
It can be seen that the performance of CoT with gold reasoning steps is much better than that of CoT, which indicates that the gap between training and inference does exist in standard CoT.

For the second observation, it might indicate that during the training of PCCoT, the student task serves as a regularizer for the teacher CoT task and helps the model to learn better reasoning paths.
We are not sure about this and further investigation is needed to understand this phenomenon.

\subsection{Visualization of the Attention Map}

\begin{figure*}[tb]
  \centering
  \begin{subfigure}{.48\textwidth}
    \centering
    \includegraphics[page=1,width=\textwidth,clip]{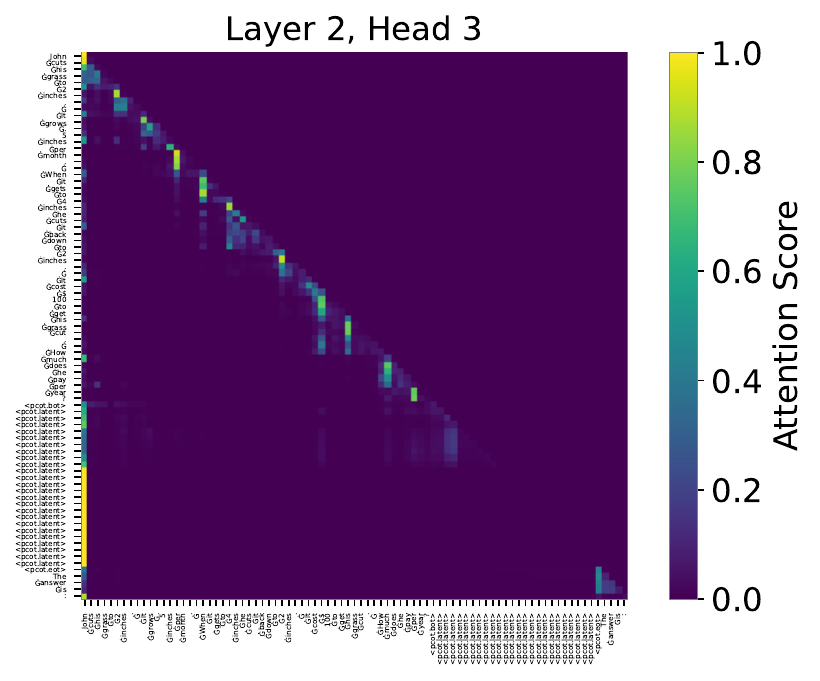}
    \caption{}
    \label{fig:attn-l2h3}
  \end{subfigure}
  \begin{subfigure}{.48\textwidth}
    \centering
    \includegraphics[page=1,width=\textwidth,clip]{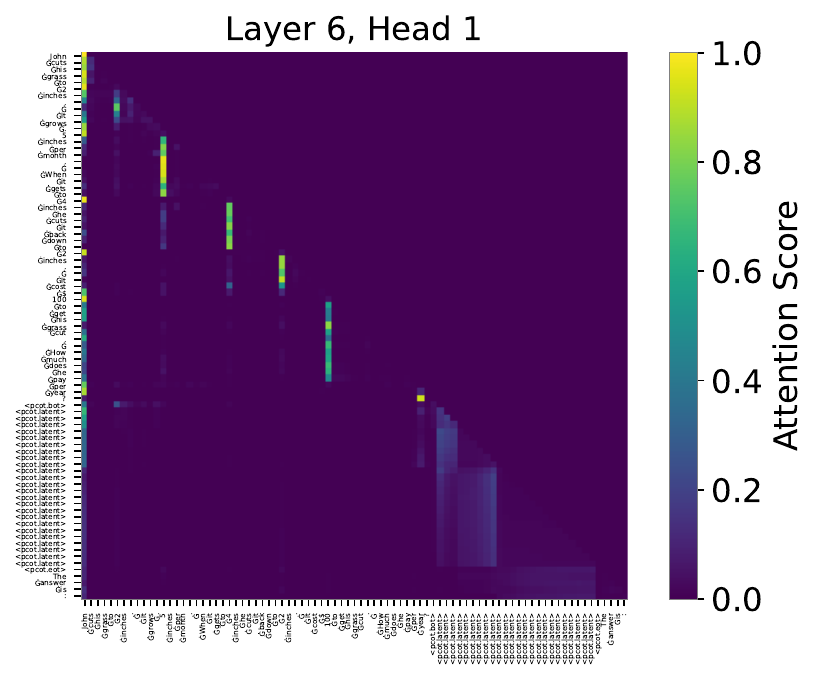}
    \caption{}
    \label{fig:attn-l6h1}
  \end{subfigure}
  \caption{The attention map in PCCoT with $c=24$ latent thought tokens and $T=3$ extra iterations. The model is finetuned from GPT-2 Small on GSM8K-Aug. The input question is ``John cuts his grass to 2 inches.  It grows .5 inches per month.  When it gets to 4 inches he cuts it back down to 2 inches.  It cost \$100 to get his grass cut.  How much does he pay per year?''. It is taken from the dev set of GSM8K-Aug.}
  \label{fig:attn}
\end{figure*}

We visualize the attention map of PCCoT to understand how the latent thought tokens interact with other tokens. Figure~\ref{fig:attn} shows the attention map of two different heads at different layers of PCCoT with $c=24$ latent thought tokens and $T=3$ extra iterations. In Figure~\ref{fig:attn-l2h3}, the latent thought tokens put most of the attention at the sink token \cite{xiao2024efficient}, especially the last 15 latent thought tokens. The answer prompts attend to the end-of-thought token and previous answer prompt tokens. The final answer prompt token mostly attends to the sink token. In Figure~\ref{fig:attn-l6h1}, the latent thought tokens evenly put attention on some previous latent thought tokens. The answer prompts also attend to the latent thought tokens.

Interestingly, we do not find any special latent thought tokens in the attention map: there does not exist a latent token that has significantly different attention patterns from others. It may indicate that the latent thought tokens carry the information evenly and this process is significantly different from the standard CoT, where the generation of the answer tokens is heavily dependent on some specific important reasoning tokens \cite{lindsey2025biology}.

\subsection{Chat Templates}

In Section~\ref{sec:exp-setup}, we mention that we use Llama3.2-1B-Instruct as one of the models in our experiments. Llama3.2-1B-Instruct is a chat model, but to fairly compare it with GPT-2 Small, we do not use the chat template during training and inference.

\begin{table}[tb]
  \centering
  \begin{tabular}{@{}llc@{}}
  \toprule
  Setting           & Dataset  & Accuracy \\ \midrule
  w/o chat template & GSM8K    & 53.35 {\small $\pm 0.18$} \\
  w/ chat template  & GSM8K    & 52.24 {\small $\pm 1.14$} \\
  w/o chat template & GSM8K-NL & 50.72 {\small $\pm 1.39$} \\
  w/ chat template  & GSM8K-NL & 51.76 {\small $\pm 0.37$} \\ \bottomrule
  \end{tabular}
  \caption{Comparison of accuracy with and without chat template on GSM8K and GSM8K-NL datasets.}
  \label{tab:chat-template}
\end{table}

Table~\ref{tab:chat-template} shows the test set accuracy of PCCoT with and without chat template on GSM8K and GSM8K-NL datasets. In general, the performance of PCCoT with and without chat template does not show a significant difference. Specifically, on GSM8K, the performance of PCCoT without chat template is slightly better than that with chat template, while on GSM8K-NL, the performance of PCCoT with chat template is slightly better than that without chat template. This might because the GSM8K-NL dataset has the reasoning steps in natural language, which is more compatible with the chat template, while the GSM8K dataset has the reasoning steps in a more structured format, which might not benefit from the chat template as much.

\section{Experiment Details}
\label{apx:exp}

In this section, we provide more details about the experiments and some justifications for the hyperparameter choices.

\subsection{Model and Training Details}

\begin{table}[tb]
  \centering
  \begin{tabular}{@{}lc@{}}
  \toprule
  Hyperparameter     & Value     \\ \midrule
  LoRA rank $r$      & 128       \\
  LoRA $\alpha$      & 32        \\
  LoRA dropout       & 0.1       \\
  LoRA bias          & False     \\
  LoRA target module & attn, ffn \\ \bottomrule
  \end{tabular}
  \caption{LoRA configurations.}
  \label{tab:lora-detail}
\end{table}

\begin{table*}[tb]
  \centering
  \begin{tabular}{@{}lcccc@{}}
  \toprule
  Model             & \multicolumn{2}{c}{GPT-2 Small} & \multicolumn{2}{c}{Llama3.2-1B-Instruct} \\ \cmidrule(l){2-5} 
  Dataset           & GSM8K         & GSM8K-NL        & GSM8K             & GSM8K-NL             \\ \midrule
  CODI $\alpha$     & 1             & 1               & 1                 & 1                    \\
  CODI $\beta$      & 1             & 1               & 1                 & 1                    \\
  CODI $\gamma$     & 1             & 1               & 20                & 20                   \\
  learning rate     & 3e-3          & 3e-3            & 8e-4              & 8e-4                 \\
  lr scheduler      & \multicolumn{4}{c}{cosine}                                                 \\
  optimizer         & \multicolumn{4}{c}{AdamW}                                                  \\
  $\beta_1$         & \multicolumn{4}{c}{0.9}                                                    \\
  $\beta_2$         & \multicolumn{4}{c}{0.999}                                                  \\
  batch size        & \multicolumn{4}{c}{128}                                                    \\
  warmup ratio      & \multicolumn{4}{c}{0.03}                                                   \\
  weight decay      & 1e-2          & 1e-2            & 1e-1              & 1e-1                 \\
  gradient clipping & 1.0           & 1.0             & 1.0               & 1.0                  \\
  epochs            & 40            & 40              & 10                & 10                   \\
  GPU               & H800x2        & A100x2          & H20x4             & H20x4                \\ \bottomrule
  \end{tabular}
  \caption{Training details of different models. The $\alpha$, $\beta$ and $\gamma$ are the hyperparameters in CODI \cite{shen2025codicompressingchainofthoughtcontinuous}. The batch size is the total effective batch size across all GPUs.}
  \label{tab:training-detail}
\end{table*}

We provide the LoRA configurations and training details in Table~\ref{tab:lora-detail} and \ref{tab:training-detail}.
Most configurations are consistent with those in CODI \cite{shen2025codicompressingchainofthoughtcontinuous}.
During training, only the embeddings of the added special tokens (\texttt{<bot>}, \texttt{<latent>}, \texttt{<eot>}) are trainable.
The embeddings of all other tokens are freezed.
We use GSM8K-Aug and GSM8K-Aug-NL \cite{deng2023implicitchainthoughtreasoning} as our datasets, which are licensed under MIT.
Our use of the datasets is consistent with their intended use.
There are 385,620 training samples in each dataset with a validation set of 500 samples and a test set of 1,319 samples.
When autoregressively decoding tokens, we always use the greedy decoding strategy.
For the three random runs, we use 0, 1, 2 as the random seeds respectively.
Our implementation is based on HuggingFace Transformers \cite{wolf-etal-2020-transformers} with kernel replacement with FlashAttention 2 \cite{dao2023flashattention2}.

\subsection{Justifications for Hyperparameter Choices}

\paragraph{LoRA instead of Full Finetuning}
We use LoRA instead of full finetuning since we find our model would easily overfit under full finetuning. The dev loss will increase quickly after a certain amount of training.

\paragraph{Not using MLP for Latent Thought Tokens}
CODI \cite{shen2025codicompressingchainofthoughtcontinuous} adds an additional trainable MLP followed by layer norm to transform the final hidden representations of the latent thought tokens before feeding them into the next step.
This introduces additional parameters and breaks the fair comparison to the baseline.
Moreover, our experiments show that adding an additional MLP on PCCoT has negligible improvement on the performance.
This is consistent with CODI, where the additional MLP only improves 1.2\% of accuracy.
Therefore, we decide not to use MLP for latent thought tokens in CODI for our experiments.

\paragraph{Greedy Decoding instead of Sampling}
We empirically find that the model performance has negligible difference when using different decoding strategies. This may because the questions on GSM8K are relatively short and easy.

\paragraph{Inference Time Measurement}
In Table~\ref{tab:latency}, we measure the inference time of different approaches with a batch size of 100 and only the time for processing the question and CoT tokens is included.
We do not include the time for generating the answer tokens since the number of answer tokens may differ between different approaches.
Additionally, the number of gold answer tokens is usually only 1 or 2, which is negligible compared to the number of question and CoT tokens.
The 100 samples are selected from the test set of GSM8K with an average length of 105 tokens.

\paragraph{Settings of Baseline Approaches}
In Section~\ref{sec:exp-setup}, we mention the settings of the baseline approaches.
We choose the best-performing setting for each approach.
Specifically, Pause Tokens uses 24 trainable pause tokens, continuous CoT uses 12 latent thought tokens, and PCCoT uses $c=24$ latent thought tokens with $T=3$ extra iterations.
Though CODI \cite{shen2025codicompressingchainofthoughtcontinuous} reports 6 latent thought tokens is the best performing setting for continuous CoT, we find that using 12 latent thought tokens is much better than using 6 in our experiments.
Notice that the average number of CoT tokens in GSM8K-Aug is 20.32, the choice of $c=24$ latent thought tokens somewhat indicates a replacement of explicit reasoning tokens with implicit reasoning tokens.

\paragraph{Choice of the Datasets}
We choose GSM8K as our dataset since they are commonly used in the literature of continuous CoT \cite{hao2024traininglargelanguagemodels,shen2025codicompressingchainofthoughtcontinuous, deng2024explicitcotimplicitcot,cheng2024compressedchainthoughtefficient}.
We have also tried other datasets, including CLUTRR \cite{sinha-etal-2019-clutrr}, CommonsenseQA \cite{talmor-etal-2019-commonsenseqa} and StrategyQA \cite{geva-etal-2021-aristotle}.
However, we cannot distinguish the performance of different approaches on these datasets, which may be due to the fact that these datasets have limited training samples.



\section{Data Examples}

In this section, we provide some data examples from the test set of the datasets we used in our experiments.

\subsection{GSM8K-Aug}

\paragraph{Question}

\begin{verbatim}
Every day, Wendi feeds each of her
chickens three cups of mixed chicken
feed, containing seeds, mealworms and
vegetables to help keep them healthy.
She gives the chickens their feed in
three separate meals. In the morning,
she gives her flock of chickens 15 cups
of feed.  In the afternoon, she gives
her chickens another 25 cups of feed.
How many cups of feed does she need to
give her chickens in the final meal of
the day if the size of Wendi's flock
is 20 chickens?
\end{verbatim}

\paragraph{Step 1}

\begin{verbatim}
<<3*20=60>>
\end{verbatim}

\paragraph{Step 2}

\begin{verbatim}
<<60-15-25=20>>
\end{verbatim}

\paragraph{Answer}

\begin{verbatim}
20
\end{verbatim}

\subsection{GSM8K-Aug-NL}

\paragraph{Question}

\begin{verbatim}
Every day, Wendi feeds each of her
chickens three cups of mixed chicken
feed, containing seeds, mealworms and
vegetables to help keep them healthy.
She gives the chickens their feed in
three separate meals. In the morning,
she gives her flock of chickens 15 cups
of feed.  In the afternoon, she gives
her chickens another 25 cups of feed.
How many cups of feed does she need to
give her chickens in the final meal of
the day if the size of Wendi's flock
is 20 chickens?
\end{verbatim}

\paragraph{Step 1}

\begin{verbatim}
If each chicken eats 3 cups of feed
per day, then for 20 chickens they
would need 3*20=60 cups of feed per
day.
\end{verbatim}

\paragraph{Step 2}

\begin{verbatim}
If she feeds the flock 15 cups of
feed in the morning, and 25 cups in
the afternoon, then the final meal
would require 60-15-25=20 cups of
chicken feed.
\end{verbatim}

\paragraph{Answer}

\begin{verbatim}
20
\end{verbatim}










\end{document}